\documentclass[letterpaper, 10 pt, conference]{ieeeconf}  

\usepackage{graphicx}
\usepackage{dblfloatfix}
\usepackage{hyperref}
\usepackage{tabularx}
\usepackage{fancyhdr}

\IEEEoverridecommandlockouts                              

\usepackage{xcolor}
\usepackage{color,soul}
\usepackage{comment}
\usepackage{pifont}
\usepackage{amsmath}

\def\baselinestretch{0.985}

\title{\LARGE \bf
SimNet: Learning Reactive Self-driving Simulations\\from Real-world Observations
}

\author{Luca Bergamini$^*$, Yawei Ye$^*$, Oliver Scheel, Long Chen,\\Chih Hu, Luca Del Pero, Błażej Osiński, Hugo Grimmett and Peter Ondruska$^+$
\thanks{$^*$ Equal contribution.}
\thanks{$^+$ Authors are with Lyft Level 5 self-driving division. Contact: pondruska@lyft.com.}
\thanks{Data, code and videos are available at \url{simulation.l5kit.org}}
}

\begin{document}

\renewcommand{\headrulewidth}{0pt}
\fancyhf{}
\fancyhead[C]{Published at 2021 International Conference on Robotics and Automation (ICRA2021)}

\maketitle
\thispagestyle{fancy}
\pagestyle{fancy}

\begin{abstract}
In this work we present a simple end-to-end trainable machine learning system capable of realistically simulating driving experiences. This can be used for verification of self-driving system performance without relying on expensive and time-consuming road testing. In particular, we frame the simulation problem as a Markov Process, leveraging deep neural networks to model both state distribution and transition function. These are trainable directly from the existing raw observations without the need of any handcrafting in the form of plant or kinematic models. All that is needed is a dataset of historical traffic episodes. Our formulation allows the system to construct never seen scenes that unfold realistically reacting to the self-driving car's behaviour. We train our system directly from 1,000 hours of driving logs and measure both realism, reactivity of the simulation as the two key properties of the simulation. At the same time we apply the method to evaluate performance of a recently proposed state-of-the-art ML planning system \cite{bansal2018planning-5} trained from human driving logs. We discover this planning system is prone to previously unreported causal confusion issues that are difficult to test by non-reactive simulation. To the best of our knowledge, this is the first work that directly merges highly realistic data-driven simulations with a closed loop evaluation for self-driving vehicles. We make the data, code, and pre-trained models publicly available to further stimulate simulation development.
\end{abstract}

\section{INTRODUCTION}

Self-Driving Vehicles (SDVs) have the potential to radically transform society in the form of safe and efficient transportation. Modern machine learning methods have enabled much of the recent advances in self-driving perception \cite{Chen_2017_CVPR,zhou2017voxelnet,qi2018frustumnet,ku2018perception3,Lang_2019_CVPR}, prediction \cite{chai2020multipath,lee2017desire,Rhinehart_2019_ICCV,tang2019multiple} and planning~\cite{bansal2018planning-5, zeng2019end}. The availability of large datasets unlocked significantly higher performance compared to older, hand-engineered systems.

However, the problem of validating SDV performance remains still largely unsolved. Most industry players validate empirically by deploying their self driving systems to a fleet of vehicles accompanied by safety drivers. In the case of unusual or failure behaviours, the safety driver takes over. Observed issues serve as feedback to improve the system. However, this process is both expensive and time-consuming, requiring the collection of thousands or even millions of miles, depending on the system's maturity. It is also hard to replicate or directly compare the performance of different system versions, as it is impossible to experience exactly the same driving situation twice.

\begin{figure}
    \centering
    \includegraphics[width=\linewidth]{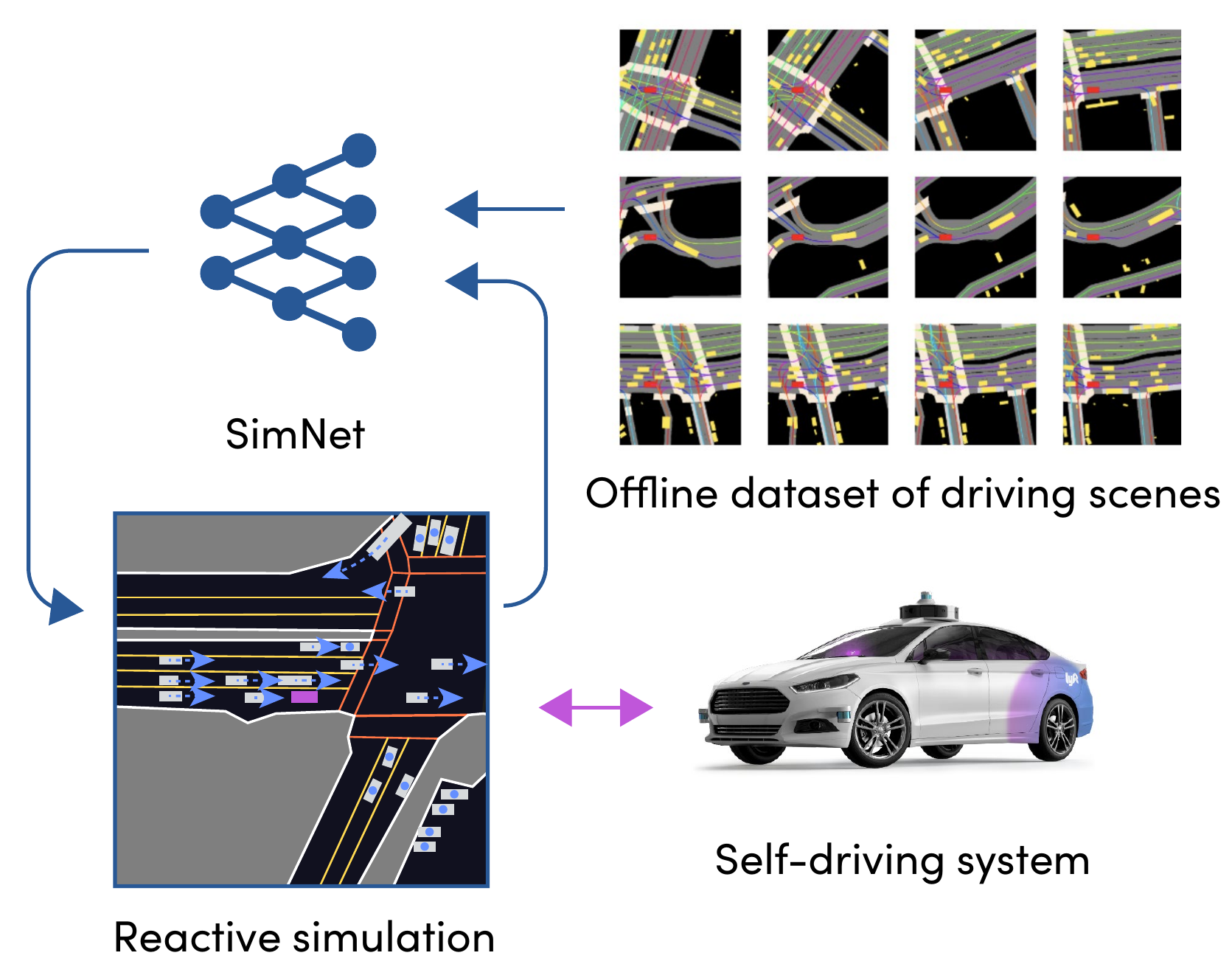}
    \caption{The proposed trainable simulation system. We frame the simulation problem as reactive episode synthesis that can be used to validate the performance of a self-driving driving system.}
    \label{fig:intro}
\end{figure}

A common approach to mitigate some of these issues is \emph{log replay}, where the movement of other traffic participants is replayed around the SDV in simulation as it happened when the log was collected. However, if the SDV's new actions differ from those when the log was collected, the traffic participants don't react to it, and thus the simulation becomes unrealistic and ineffective for validation. For example, even a slight braking during the log replay can result in an unrealistic collision with the trailing car due to non-reactivity. These unrealistic outcomes are a result of what is called \emph{simulation drift}.

One way to implement simulation reactivity is by scripting traffic participant behaviour to follow certain rules. However, this is time consuming and still lacks the realism and fidelity of road testing, thus undermining the validation effort.

In this paper, we aim to create realistic simulated driving experiences just like those that an SDV would encounter in the real world. For example, when the SDV decides to slow down, the simulated vehicle behind it should react by either slowing or overtaking, just as it would in a road test. Additionally, agent behaviour should capture the stochastic multi-modality that we observe on the road.

To achieve this, we frame simulation as an ML problem, in which we generate driving episodes that need to be both \emph{realistic} and \emph{reactive} to SDV behaviour. We then present a system leveraging high-capacity ML models trained on large amounts of historical driving data, Figure \ref{fig:intro}. We show the system's performance dramatically improves as the amount of data grows significantly narrowing the gap between road testing and offline simulation. At the same time, we show that it can help to identify issues in state-of-the-art planning systems. To the best of our knowledge, this is the first work that connects simulation and planning with large-scale datasets in a realistic self-driving setting.

Our contributions are four-fold:
\begin{enumerate}
    \item The formulation of the self-driving simulation problem as an ML problem which seeks to generate driving episodes that are both realistic and reactive to SDV behaviour;
    \item A simple machine-learned simulation system that can sample these episodes based on historical driving data trainable directly from traffic observations;
    \item Qualitative and quantitative evaluation of the proposed method as a function of training data size, as well as, its usefulness in evaluating and discovering issues in a state-of-the-art ML planning systems.
    \item The code and the pretrained models of the experiments to further stimulate development in the community.
\end{enumerate}

\section{RELATED WORKS}
Our work is situated in the broader context of trajectory prediction, planning and simulation for autonomous vehicles.

Ability to predict future motion of traffic participants around the vehicle is important to be able to anticipate future necessary for planning. Classical methods include dynamic models~\cite{brannstrom2010dynamic-1,lin2000dynamic-2,huang2006dynamic-3}, kinematic models~\cite{ammoun2009kinematic-1,hillenbrand2006kinematic-2,miller2002kinematic-3}, Kalman filter-based systems~\cite{dyckmanns2011kalman-1,veeraraghavan2006kalman-2}, Monte Carlo sampling~\cite{broadhurst2005monte-1,eidehall2008monte-2,althoff2011monte-3} and trajectory prototypes~\cite{hermes2009pattern-1,hu2006pattern-2,atev2010pattern-3}.
Today, deep learning methods for trajectory predictions are widely adopted.
Notable examples include \cite{chai2020multipath}, \cite{fang2020tpnet}, and \cite{liang2020learning}.
Authors of~\cite{chai2020multipath} leverage bird's-eye view (BEV) rasters and a fixed set of future trajectory anchors. During training, the model learns displacement coefficients from those anchors along with uncertainties.
TPNet~\cite{fang2020tpnet} is a two stage network, where during the first stage the final future waypoint of the trajectory is predicted, starting from BEV semantic rasters. Then, proposals are generated to link past observations with this final waypoint and points near to it. We take inspiration from the above methods leveraging deep neural network architectures but intend to solve motion simulation instead of one-shot motion prediction. The key difference is that in motion simulation the sequence of traffic agent motion is generated one timestep at a time reflecting on both the intention of traffic agents themselves but also motion of other traffic participants and SDV.

Given the prediction of future motion of traffic participants SDV plans its own actions. Historically, this has been tackled by various methods optimising an expert cost function \cite{buehler2009darpa, fan2018baidu, ziegler2014trajectory}. This cost function captures various desirable properties i.e. distance to other cars, comfort of the ride, obeying traffic rules etc. Engineering this cost function is, however, complex and time-consuming. Recently, novel methods \cite{bansal2018planning-5, zeng2019end, wulfmeier2017large} were proposed that frame the problem of planning as learning from demonstrations. Authors from~\cite{behbahani2019learning} use an inverse reinforcement learning technique to learn from expert demonstrations. \cite{henaff2018model} deals with accumulating errors and presents a method trained entirely from data. A limitation of this model is that it directly predicts the visual output, which results in unnecessary errors, such as a car changing shape in consecutive frames. 

In our work we do not try to build a planning system but to accurately evaluate performance of an existing one. In particular, we aim to explore performance of a recently proposed ML planning system \cite{bansal2018planning-5}. This is a particularly appealing application given the novelty of ML planning systems in self-driving and their relatively unexplored performance. We find out that this particular method, while promising, is prone to causal confusion of imitation learning \cite{de2019causal} that incorrectly associates motion of other cars with the desired action of the SDV. This issue is difficult to observe in non-reactive situation but becomes apparent when using a realistic reactive simulation.


Another way to perform simulation is to use an advanced driving simulator with agents controlled by hand-crafted rules. Notable examples of such driving simulators are SUMO \cite{SUMO2018} and CARLA \cite{dosovitskiy2017carla}. A disadvantage of this solution is that hand-coded actors tend to be unrealistic and rarely present a wide enough variety of behaviours. Our aim is to achieve high realism by directly learning behaviour of other traffic participants from observed data. We show that these methods can be very powerful and their accuracy improves significantly with the amount of data used for training. As collecting these data is significantly easier than engineering a realistic simulation, this approach is much more scalable.
 
\begin{figure*}
    \centering
   \includegraphics[width=\linewidth]{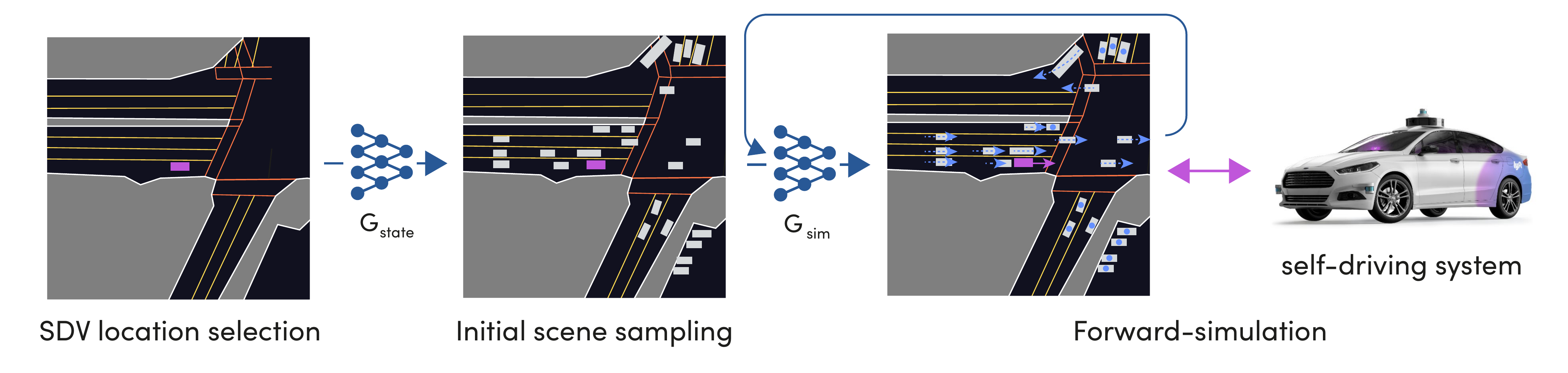}
    \caption{Overview of the proposed simulation sampling process. To generate a new driving episode we first pick and sample an initial state capturing the positions of all traffic participants. Next, the state is forward-simulated with the traffic participants controlled by a neural network and the behaviour of SDV controlled by a self-driving control loop.}
    \label{fig:system}
\end{figure*}

\section{SELF-DRIVING SIMULATION AS A LEARNING PROBLEM}
In this section, we formulate the simulation problem as a machine learning problem. Specifically, we aim to sample realistic driving experiences that the SDV would encounter in the real world, that also react to the SDV behaviour given by its control policy $f$. To help model this realism, we have access to historical driving scenes $D$ that the capture observed behaviour of other traffic participants on top of a semantic map $\mathcal{M}$ in a variety of diverse driving scenarios.

Each driving episode of length $T$ can be described as a sequence of observed states $s_1, s_2, ..., s_T$ with each state capturing the position, rotation, size and speed of all nearby traffic participants $z$:
\begin{equation}
    s_t = \{z^1_t, z^2_t, ..., z^k_t\}.
\end{equation}
This representation corresponds exactly to the output of the perception system, which turns raw sensor measurements into the vectorised detections of traffic participants and can then be fed to the SDV's control algorithm.


The SDV itself is modelled simply as one of these participants $z^{\text{SDV}}$, but unlike other traffic participants its dynamics are controlled by a known function $f$ implementing the self-driving algorithm:
\begin{equation}
    z^{\text{SDV}}_{t+1} = f(z^{\text{SDV}}_t, s_t).
\end{equation}
Generating new driving experiences can then be described as sampling from the joint distribution of the stochastic behaviour of other traffic participants and the deterministic SDV behaviour.

Note that this joint distribution is inherently non-deterministic. Given an initial state $s_1$ there are many possible ways that the future can unfold. At each moment in time, traffic participants must act and react to new information, such as attempts to merge, nudge, slow down, resulting in a complex set of driving behaviours. In the following section, we describe a method leveraging deep learning that can realistically sample such reactive episodes.

At the same time we aim to compute a certain set of metrics that describe the performance of the self-driving system, such as, the amount of \textit{collisions}, \textit{traffic rules violations} etc. An important property is the accuracy of these metrics or a ratio of false positive vs. false negative events. In an ideal simulation the amount of both is close to 0 but any deeper understanding about the correct attribution is valuable.

\begin{figure*}
    \centering
    \includegraphics[width=\linewidth]{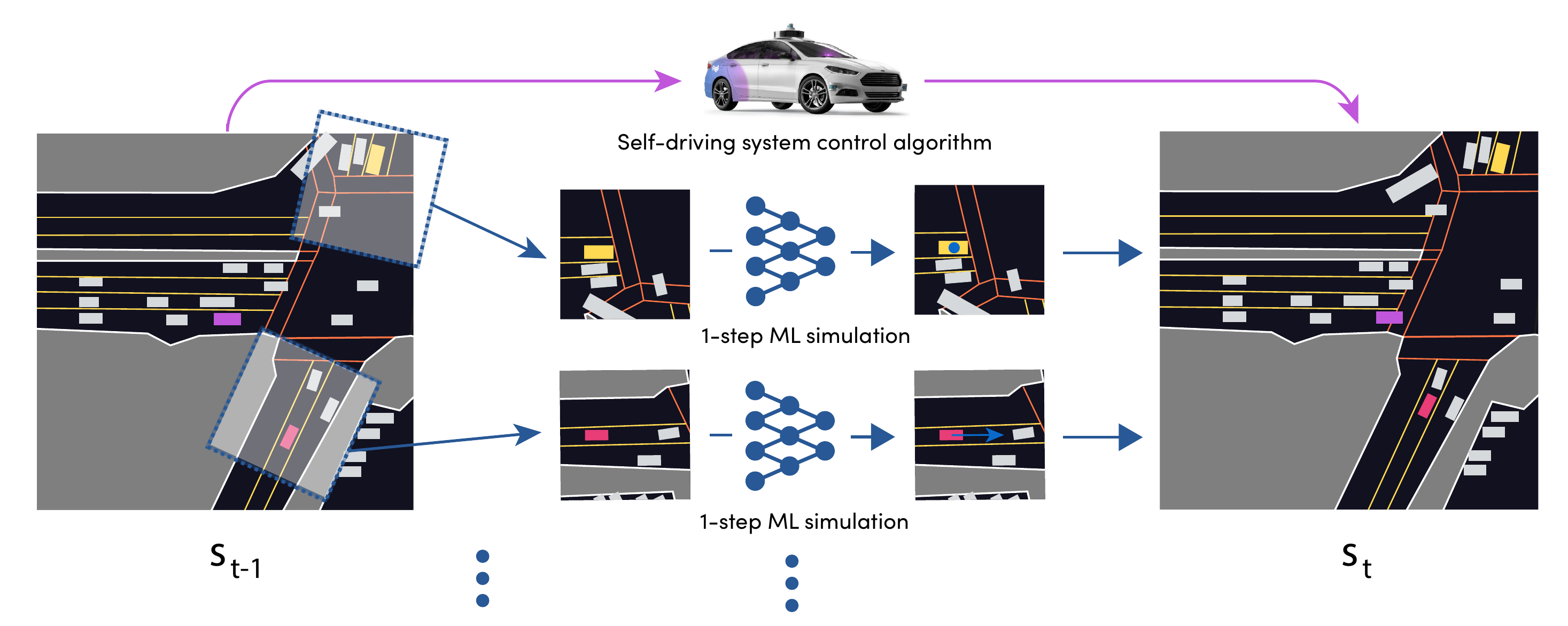}
    \caption{Detail of the interactive state unroll. For all agents in the state $s_{t-1}$ we independently run 1-step prediction to advance them. The self-driving car is controlled by control algorithm $f$. The new positions then form a new state $s_t$ and the process repeats. }
    \label{fig:unroll}
\end{figure*}

\section{GENERATING REALISTIC AND REACTIVE DRIVING EPISODES}
In this section we describe an effective way to draw realistic and reactive driving episodes as defined in the previous section.

The sampling of driving episodes can then be formalised as a Markov Process with resulting probability distribution factorising as:
\begin{equation}
    p(s_1, s_2, ..., s_T) = p(s_1) \prod^T_{t=2} p(s_t | s_{t-1}).
\end{equation}

Furthermore, we assume the actions are locally independent for each participant conditioned on each previous state:
\begin{equation}
    p(s_t | s_{t-1}) = \prod^K_{k=1} p(z^k_t | s_{t-1}).
\end{equation}
This follows the intuition of real-world driving where participants act independently, each controlling their own behaviour, observing others and reacting to new information that becomes available after each time-step.

Both the initial state distribution $p(s_1)$ and participant policy $p(z^k_t | s_{t-1})$ are modelled by a separate neural networks $G_{\text{state}}$ and $G_{\text{sim}}$. In particular, the participant transition policy is controlled by a neural network producing steering $\phi$ and velocity $v$
\begin{equation}
    \phi^k, v^k \leftarrow G_{\text{sim}}(z^k_{t-1}, s_{t-1})
\end{equation}
that are used to update particular position of participant $z^k$.

Sampling from this process consists of executing three steps as summarised in Figure \ref{fig:system}, and outlined in detail in the next subsections:
\begin{enumerate}
    \item Initial SDV location $l$ is chosen from all permissible locations on the map;
    \item Initial state $s_1$ is drawn from the distribution of all feasible states. This state captures the total number and initial poses of all traffic participants;
    \item A driving episode $s_2, ... s_T$ is generated via step-by-step forward simulation employing the participant's policy $p(z^k_t | s_{t-1})$ and self-driving control system $f$.
\end{enumerate}

This formulation offers a high degree of flexibility, allowing one to tailor the properties of the resulting simulation:
\begin{itemize}
    \item \textbf{Full simulation:} Executing all above steps results in generating new, never-experienced driving episodes from all locations.
    \item \textbf{Journey simulation:} By keeping the initial SDV location $l$ fixed, we can synthesise many different initial conditions and driving episodes starting at that position.
    \item \textbf{Scenario simulation:} By using an existing historical state of interest as $s_1$, we can generate many resulting possible futures.
    \item \textbf{Behaviour simulation:} We can replace steering angle $\phi$ by hard-coding a specific path for them to follow. This forces a particular high-level behaviour of a traffic participant but still leaves a degree of reactiveness in execution. This is useful for simulating SDV behaviour in specific situations, e.g. being cut-off by another car.
\end{itemize}

\subsection{Initial state sampling}


To represent the state $s$ around the self-driving vehicle, we leverage a bird's-eye view representation rendering positions of nearby traffic participants on top of a semantic map $\mathcal{M}$. This representation has proven to be an effective representation in recent motion prediction and planning works \cite{cui2019multimodal, bansal2018planning-5}. One advantage is that it effectively captures both local context and a variable amount of traffic participants in the form of a single image $I_s$.

To sample an initial state $s_1$ similar to our training distribution, we leverage conditional generative adversarial networks (cGANs) \cite{isola2017image} conditioned on empty scenes capturing only the semantic map $I_\mathcal{M}$. This network is trained on pairs of $\{I_{\mathcal{M}}, I_{s}\}$ constituting the training dataset. Specifically, the generator network is trained to convert $I_\mathcal{M}$ into $I_{s}$ and the discriminator to distinguish the synthetic states $I_{s}$ from real ones. Upon convergence, the generator can create unlimited amounts of new synthesised states $s_1$ for any map location, indistinguishable from real ones, to seed the simulation.

To extract the final numerical positions and rotations of the vehicles $z_1, z_2, ... z_K \in s$, we use a connected components algorithm. For each connected component we compute the centroid and a minimum bounding box capturing its size.


\subsection{Forward simulation}
This step generates the full sequence $s_2, s_3, ..., s_T$. This generation happens one step at a time, executing policy $p(z^k_t | s_{t-1})$ for each traffic participant and SDV control policy $f$ for the self-driving vehicle. 

As the traffic participant policy we employ a model described in \cite{cui2019multimodal} consisting of a ResNet-50 backbone that takes bird's-eye-view rasterised states $s_t$ around the traffic participants $z^k_t$ as input, and a regression head predicting steering $\tau^k_t$ and velocity $v^k_t$. We train the model on past agent trajectories. In particular, we take all traffic participants from the training dataset $D$ with observation history longer than 1s and train the model to predict their individual steering and velocity.

As illustrated in Figure \ref{fig:unroll}, to compute a new state $s_t$ the policy is invoked for every observed traffic participant $z^t_k$ in the current state independently. Simultaneously, vehicle controls are triggered to obtain the SDV's new position:
\begin{equation}
    z^{\text{SDV}}_t = f(s_{t-1}, z^{\text{SDV}}_{t-1}).
\end{equation}
This then constitutes a new state $s_t$, and the process repeats until the entire episode is generated or simulation is interrupted, i.e. due to a simulated SDV collision.

\begin{figure}[t]
    \centering
    \includegraphics[width=\linewidth]{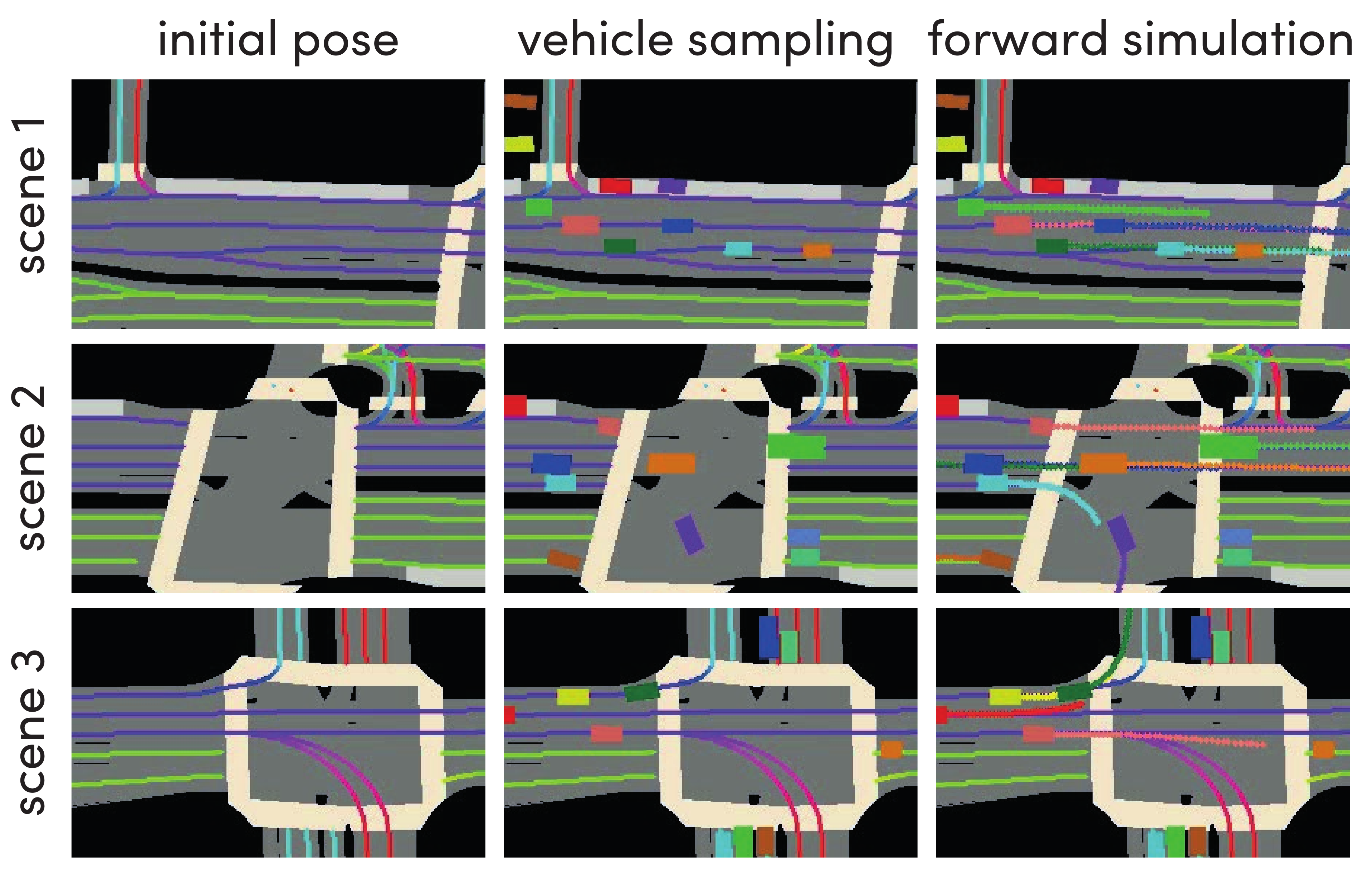}
    \caption{An example of initial state and intended traffic participant trajectories. Each row shows a separate exemplar episode. From left to right: the initial scene sampled from SDV positions with agents being removed, the sampled traffic participants' positions, and the trajectory taken by each vehicle.}
    \label{fig:initial}
\end{figure}

\section{EXPERIMENTS}

\begin{figure}[t]
    \centering
    \includegraphics[width=\linewidth]{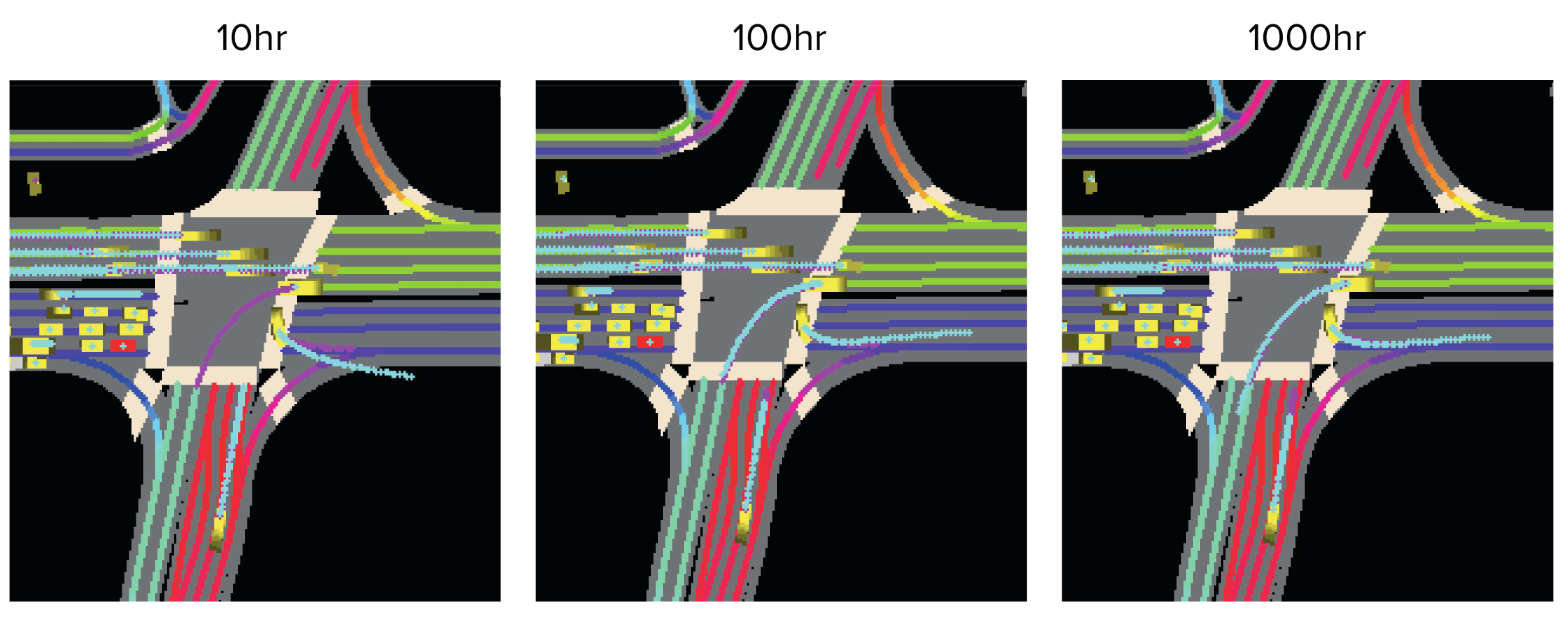}
    \caption{Qualitative example of the simulation trained on various amount of data. More training data leads to more realistic simulations, with model predicted paths shown as blue lines and ground truth paths as purple lines.}
\label{fig:perf_over_data}
\end{figure}

\begin{table}[t]
\centering
\begin{tabular}{l|llllll|l}
\multicolumn{1}{c}{} & \multicolumn{6}{c}{\textbf{Displacement error [m]}} & \multicolumn{1}{c}{}\\
\textbf{Method} & 0.5s  & 1s & 2s  & 3s & 4s & 5s & \textbf{Reactivity}  \\ \hline
Log replay & 0 & 0 & 0 & 0 & 0 & 0 & 0 \\
10 hr   & 0.58 & 0.95 & 1.76 & 2.57 & 3.40 & 4.28 & 0.95\\
100 hr  & 0.51 & 0.81 & 1.44 & 2.04 & 2.61 & 3.19 & \textbf{0.97}\\
1000 hr & \textbf{0.49} & \textbf{0.76} & \textbf{1.32} & \textbf{1.87} & \textbf{2.42} & \textbf{2.99} & \textbf{0.97}
\end{tabular}
\caption{Realism and reactiveness of non-reactive and reactive simulation trained on various amounts of data for various simulation horizons. Log-replay is perfectly realistic but not reactive while SimNet is both realistic and reactive.}
\label{tab:results}
\vspace{-5mm}
\end{table}


Here we provide qualitative and quantitative evaluation of the proposed simulation system. In particular, we are interested in the system's ability to synthesise realistic initial states, forward-simulate full driving episodes, and to react to the SDV's behaviour. To evaluate it, we forward-simulate for 5 seconds across many scenes. Unrolling for 5 seconds is enough to capture interesting maneuver while still being able to collect ground truth annotations from the tracked agents in the dataset.

We capture the system's performance using two metrics. Both metrics are evaluated on 960 scenes, although these sets of scenes are disjoint.
 
\textbf{Simulation realism:} An average L2 distance between simulated agents and their ground-truth positions at different time steps into the future (displacement). For this experiment we initialise the state from a real-world log, and the SDV follows exactly the same path as it did in that log. A perfect simulation system should be able to replicate the behaviour of other agents as it happens in the log.

\textbf{Simulation reactivity:} We measure collision rate in synthetic scenarios where a static car is placed in front of a moving car, see the first two rows in Figure~\ref{fig:log_replay_simnet_combined}. This should not cause collisions in reality, and requires trailing cars to react by stopping. We report the number of scenes without a collision divided by the total number of scenes tested.

\subsection{Implementation Details}
\label{sec:implementation_details}
We train and test our approach on the recently released Lyft Motion Prediction Dataset~\cite{houston2020one}. The dataset consists of more than 1,000 hours of dense traffic episodes captured from 20 self-driving vehicles. We follow the proposed train / validation / test split. We rasterise the high-definition semantic map included in the dataset to create bird's-eye view representations of the state, centered around each agent of interest to predict its future trajectories. This representation includes lanes, crosswalks and traffic light information. At crossings, lanes are not rendered if the controlling traffic light is red. As for agents, we focus our attention on vehicles~(92.47\% of the total annotated agents), pedestrians~(5.91\%) and cyclist~(1.62\%).
We use rasters of size 224x224, a batch size of 64 and the Adam~\cite{kingma2014adam} optimizer in all our experiments. During evaluation, the whole pipeline takes around 400ms per frame with a modern GPU, which is acceptable for a non real-time constrained system.

\begin{table}[b]
\centering
\begin{tabular}{@{}p{45mm}|p{15mm}|p{15mm}@{}}
\textbf{Planning metric} & \textbf{Log-replay} & \textbf{SimNet} \\ \hline
Front collisions & 2 & 2 \\
Side collision & 4 & 9\\
Rear collision & 60 & 2 \\
Displacement error & 19 & 27 \\
Passiveness & 32 & 124 \\
Distance to reference trajectory & 2 & 2 \\
\end{tabular}
\caption{Planning metrics of \cite{bansal2018planning-5} when evaluated using non-reactive and reactive simulation. Non-reactive simulation reports various issues, such as, passiveness as false positive rear collisions. These can be properly identified using reactive simulation.}
\label{tab:logreplay_vs_simnet}
\end{table}


\subsection{Initial state sampling}

A qualitative example of sampling the initial state and intended trajectory for each traffic participant is shown in Figure \ref{fig:initial}. We specifically focus on intersections as these situations give an opportunity to traffic participants to make a variety of decisions. As one can see, the generated states and scenes look realistic. Furthermore, learned trajectories effectively capture the variety of possible participant behaviours.

\begin{figure*}
    \centering
    \includegraphics[width=\linewidth]{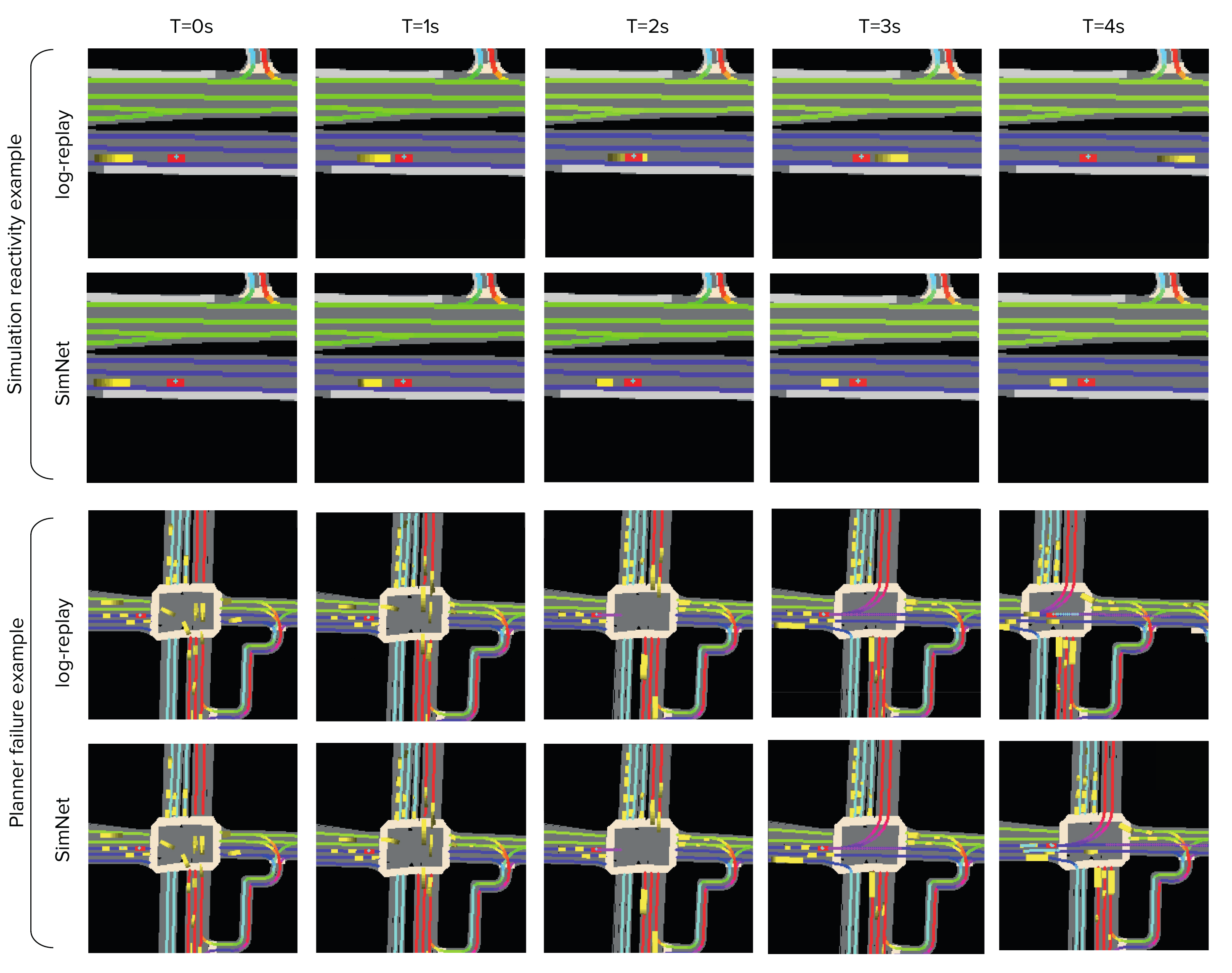}
    \caption{The first two rows: a qualitative example of the reactivity evaluation - the agent controlled by SimNet [yellow] stopped behind the static vehicle [red], while the log-reply crashed into it without showing any reactivity. Last two rows: we found the reactivity of SimNet can expose causal confusion of ML planner - SDV waits for a slight movement of the chase car to start moving as would happen in log-replay. In reactive simulation this signal does not come and the SDV keeps waiting at the intersection. Best viewed in high-resolution.}
\label{fig:log_replay_simnet_combined}
\vspace{-5mm}
\end{figure*}


\subsection{Forward simulation}

The evaluation of SimNet demonstrates very good performance with respect to both simulation realism and reactiveness. The quantitative results can be found in Table~\ref{tab:results}. The performance improves with the amount of data used for training. The qualitative comparison of models' realism is presented in Figure~\ref{fig:perf_over_data}. Similarly, simulation reactiveness is presented in Figure~\ref{fig:log_replay_simnet_combined}.

\subsection{Evaluating SDV with SimNet}


The purpose of SimNet is to accurately evaluate the performance of the SDV. In this section we describe an experiment verifying that the model is indeed fit for the purpose.  

We implemented a motion planner based on the state-of-the-art ChaufferNet \cite{bansal2018planning-5}. The model's input and backbone are the same as for SimNet (see Section~\ref{sec:implementation_details}). It predicts a future trajectory for the ego vehicle and is trained using behavioural cloning.
Naive behavioural cloning suffers from the distribution shift between training and evaluation data. Similarly to \cite{bansal2018planning-5}, we alleviate this problem by introducing synthetic perturbations to the training trajectories.

We compared two ways to evaluate such a planner, based on log-replay and SimNet. The results are presented in Table~\ref{tab:logreplay_vs_simnet}.
The number of errors turned out to differ significantly in two categories: rear collisions and passiveness.
This discrepancy raises the question of whether the reported errors reveal real mistakes of the planner or they are only artifacts of the incorrect methods of evaluation.

In order to determine this, we conducted a qualitative analysis of failure cases.
In log-replay, the cases of rear collisions consisted of both false positives (when the ego was driving slightly slower than the reference trajectory and the chasing car did not accommodate for this), as well as true positives (when the ego was passive and not starting at an intersection at a green light).
Both of these types of rear collisions disappear when evaluated in SimNet. This is to be expected, as in SimNet the chasing vehicle reacts to the static or slower ego.
This could partially explain why SimNet reports higher passiveness errors compared to log-replay.

However, the increase in passiveness (from 32 to 124) is bigger than the total number of rear collisions (60). A qualitative investigation of the scenes uncovered an interesting failure mode of the ML planner: it would not start at an intersection if neighbouring agents are static (see example in Figure~\ref{fig:log_replay_simnet_combined} below).
This is a previously unreported instance of the causal confusion \cite{dehaan2019causal}.
Moreover, it would not be possible to uncover it using log replay, because in such cases neighbouring agents (both in the front and behind the agent) will move as they did during log recording.


\section{CONCLUSIONS}
We have presented an end-to-end trainable machine learning system that generates simulations of on-road experiences for self-driving vehicles. SimNet leverages large volumes of historical driving logs to synthesize new realistic and reactive driving episodes that can be used to validate SDV performance. The evaluation shows that SimNet achieves very good results in terms of both realism and reactivity.
Moreover, using it for evaluating an ML planner has resulted in uncovering a previously unreported causal confusion of ChaufferNet \cite{bansal2018planning-5}.
Notably, we have confirmed that SimNet, which is also trained using imitation learning, exhibits the same failure mode.
We consider addressing the issue of causal confusion to be an important further work aimed at improving both ML planners and simulators.

We believe this is an exciting step towards significantly decreasing the need for on-road testing in self-driving development, and the democratisation of the field. We hope the release of our system's source code will further stimulate development in ML simulation systems.

\addtolength{\textheight}{-2cm}

\bibliographystyle{IEEEtran}
\bibliography{references}

\end{document}